\documentclass[letterpaper, 10 pt, conference]{ieeeconf}  % Comment this line out if you need a4paper

\IEEEoverridecommandlockouts                              % This command is only needed if                                                % you want to use the \thanks command
\usepackage{graphics} % for pdf, bitmapped graphics files
\usepackage{epsfig} % for postscript graphics files
\usepackage{mathptmx} % assumes new font selection scheme installed
\usepackage{times} % assumes new font selection scheme installed
\usepackage{amsmath} % assumes amsmath package installed
\usepackage{amssymb}  % assumes amsmath package installed
\usepackage{algorithm,algorithmic}
\usepackage{scalerel}
\usepackage[dvipsnames]{xcolor}

\title{\LARGE \bf
Can I lift it? Humanoid robot reasoning about the feasibility of lifting a heavy box with unknown physical properties
}

\author{Yuanfeng Han$^{*}$,  Ruixin Li$^{*}$ and  Gregory S. Chirikjian$^{\dagger,*}$% <-this % stops a space
\thanks{$^{*}$ Yuanfeng Han and Ruixin Li are with the Department of Mechanical Engineering, Johns Hopkins University, Baltimore, MD.
       {\tt\small yhan33@jhu.edu}}
\thanks{$^{\dagger,*}$ Gregory S. Chirikjian is with the Department of Mechanical Engineering,
National University of Singapore, Singapore and the Laboratory for Computational Sensing and Robotics, Johns Hopkins University, Baltimore, MD.
        {\tt\small mpegre@nus.edu.sg, gchirik1@jhu.edu}}%
}

\begin{document}

\maketitle
\thispagestyle{empty}
\pagestyle{empty}

%%%%%%%%%%%%%%%%%%%%%%%%%%%%%%%%%%%%%%%%%%%%%%%%%%%%%%%%%%%%%%%%%%%%%%%%%%%%%%%%
\begin{abstract}
A robot cannot lift up an object if it is not feasible to do so. However, in most research on robot lifting, ``feasibility" is usually presumed to exist a priori. This paper proposes a three-step method for a humanoid robot to reason about the feasibility of lifting a heavy box with physical properties that are unknown to the robot. Since feasibility of lifting is directly related to the physical properties of the box, we first discretize a range for the unknown values of parameters describing these properties and tabulate all valid optimal quasi-static lifting trajectories generated by simulations over all combinations of indices. Second, a physical-interaction-based algorithm is introduced to identify the robust gripping position and physical parameters corresponding to the box. During this process, the stability and safety of the robot are ensured. On the basis of the above two steps, a third step of mapping operation is carried out to best match the estimated parameters to the indices in the table. The matched indices are then queried to determine whether a valid trajectory exists. If so, the lifting motion is feasible; otherwise, the robot decides that the task is beyond its capability. Our method efficiently evaluates the feasibility of a lifting task through simple interactions between the robot and the box, while simultaneously obtaining the desired safe and stable trajectory. We successfully demonstrated the proposed method using a NAO humanoid robot.
\end{abstract}

%%%%%%%%%%%%%%%%%%%%%%%%%%%%%%%%%%%%%%%%%%%%%%%%%%%%%%%%%%%%%%%%%%%%%%%%%%%%%%%%
\section{Introduction}
\begin{figure}[t!]
\centering
\includegraphics[width=0.92\linewidth]{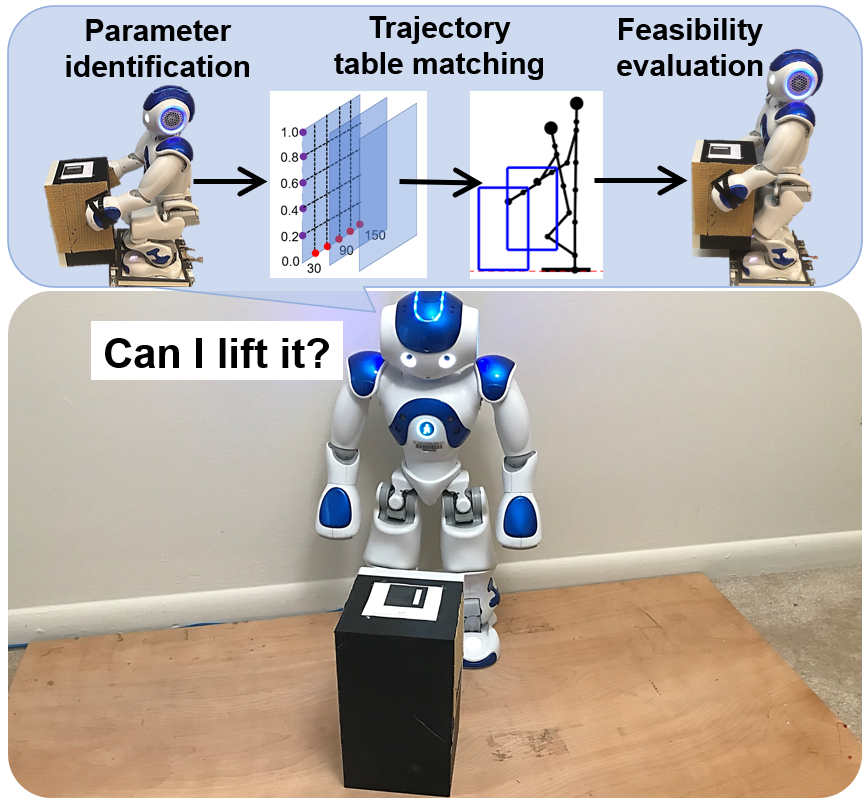}
\caption{A humanoid robot reasons about the feasibility of lifting up a heavy box with unknown physical properties using a 3-step method.}
\label{Fig.reason}
\end{figure}
Many real-world applications require humanoid robots to repeatedly manipulate certain objects with regulated or fixed sizes. Some examples include transporting packaging boxes in a warehouse \cite{yaguchi2017research} or carrying a frequently-used basket in the home \cite{edsinger2006manipulation}. Although the dimensions of such containers are often known, the inertia properties are a priori unknown due to the the varying nature of the objects that they may contain. As a result, the unpredictable nature of the inertial properties of such objects motivates the need for the robot to first reason about the feasibility of achieving such lifting tasks before committing to execution.  Here, we are specifically interested in studying this approach as applied to humanoid robots tasked with the lifting of a fixed-sized heavy box, for which the physical properties are a priori unknown. 

A humanoid robot requires a valid whole-body motion trajectory to successfully lift up a box, the existence of which is dependent upon the physical properties of the box and the capability of the robot itself. For example, the weight and the location of center of mass (COM) of the box affect not only  the stability of the robot \cite{harada2005humanoid} but also the required torque of the actuators \cite{wang2001payload,arisumi2007dynamic}. In addition, the frictional properties of the surface of the box determine whether the robot can provide enough frictional forces and torques to prevent the box from slipping from or rotating in the grip of the hands. In this regard, developing an approach to identify the above-mentioned physical parameters is crucial for further determining the feasibility of the task.  
\begin{figure*}
\centering
\includegraphics[width=1\textwidth]{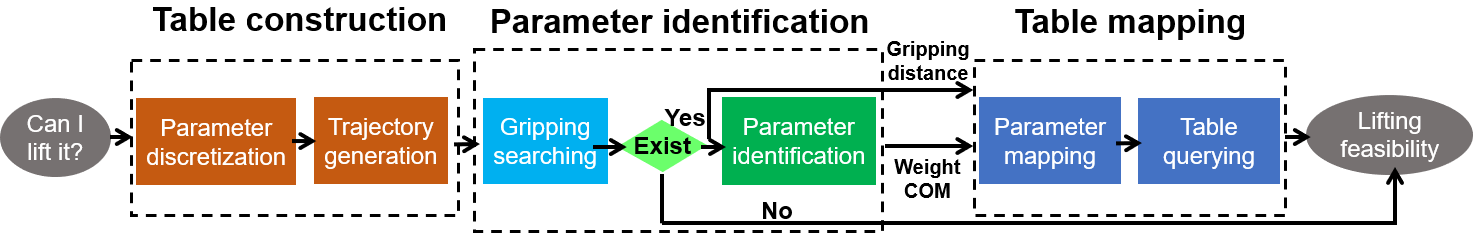}
\caption{A flow chart of the proposed reasoning method. This method includes three steps. Firstly, the physical parameters of the box are discretized and a trajectory table corresponding to these indices is constructed using simulations (left). Second, robot identifies the parameters of the box by interacting with the box (middle). Lastly, the obtained parameters are mapped to the indices of the table to evaluate the feasibility of the lifting task (right).}
\label{fig:Panel}
\end{figure*}

Many works study humanoid robots manipulating objects \cite{kemp2007challenges, asfour2008toward, zollner2004programming, ito2006dynamic,stuckler2009integrating}. Most of these are confined to the robot's ``comfortable" manipulation zone. However, many objects cannot be manipulated in practice, because they are simply too heavy for the robot to lift. Without taking the inertia of objects into account, planned motions may not be realizable, and can result in control commands that saturate actuators and can lead to damage.

Here, we propose a three-step method for humanoid robots to reason about the feasibility of lifting a box with a priori unknown physical properties (Fig. 1). The flow chart of this reasoning process is shown in Fig. 2. In the first step (Fig. 2, left), we discretize the weight, location of the COM and gripping distance (see Fig. 3A for definition) of the box within their allowable ranges. Then we generate quasi-static lifting trajectories corresponding to each combination of the above discretized indices by conducting simulations using an optimal control framework and tabulate the valid trajectories. In the second step (Fig. 2, middle), the robot identifies the physical properties of the box by implementing a physical-interaction-based planning algorithm, which searches for a robust gripping position by letting the robot repeatedly attempt to lift the box slightly off the ground while adjusting its postures. During this process, the stability of the robot is ensured and the actuator torques are limited. Finally, a third step is carried out to best match the estimated parameters with those stored in the table (Fig. 2, right). If a valid trajectory exists, the robot is considered to be able to lift up the box along an assigned trajectory; otherwise, the robot decides that the task is infeasible to execute. Our method efficiently obtains the feasibility concerning lifting a box with a priori unknown physical properties through simple interactions between the robot and the object while simultaneously determining the desired motion trajectory. We demonstrated our approach on a NAO humanoid robot.

\section{Related Work}
There exist many studies focusing on humanoid robots moving large and cumbersome objects, such as pushing heavy objects \cite{harada2003pushing}, moving wheel chairs \cite{nozawa2011full}, moving heavy boxes via a series of pivoting motions \cite{yoshida2010pivoting}, carrying a heavy object while keeping balance \cite{stephens2010dynamic}, transporting heavy objects in cluttered space \cite{rioux2015humanoid}, etc. The problem of lifting heavy objects is also mentioned in the literature. Harada et al. developed a control framework for a humanoid robot to pick up a heavy box while balancing itself \cite{harada2005humanoid}. Arisumi et al. provided a method for the dynamic lifting of a heavy box by combining two pre-designed motions \cite{arisumi2008dynamic}. Recent work from Shigematsu et al \cite{shigematsu2018lifting} presented a planning method for lifting a heavy box above the head. Prior work has provided solutions for humanoid robots to handle various challenging objects in real-world applications, for which researchers have focused on developing planning and control algorithms. However, in these works, it is known a priori that the objects can be manipulated. Very little work on reasoning the feasibility of achieving such tasks has been presented.

There are only a few studies investigating how humanoid robots can estimate the physical properties of a heavy object. In \cite{harada2005humanoid}, a humanoid robot identifies the mass and the location of the COM of a box by lifting the object up and using the force and torque sensors on the robot's wrist. This method applies only if the robot has already lifted the box off the ground and the robot's grip is of good quality. Hence, the question of whether or not the box can be lifted is not considered. In \cite{shigematsu2018lifting}, the authors propose a method for estimating both the minimal lifting friction and the mass properties of the object by letting the robot iteratively lift the box while increasing gripping force. This method first assumes the box can be lifted, and it does not consider improving the robot's gripping location to better balance the frictional torques generated by the gravity of the box. Our parameter identification method improves the above approaches by tracking robot sensory feedback from initial lifting attempts and continuously adjusting robot's postures.  
\begin{figure}[t!]
\centering
\includegraphics[width=0.95\linewidth]{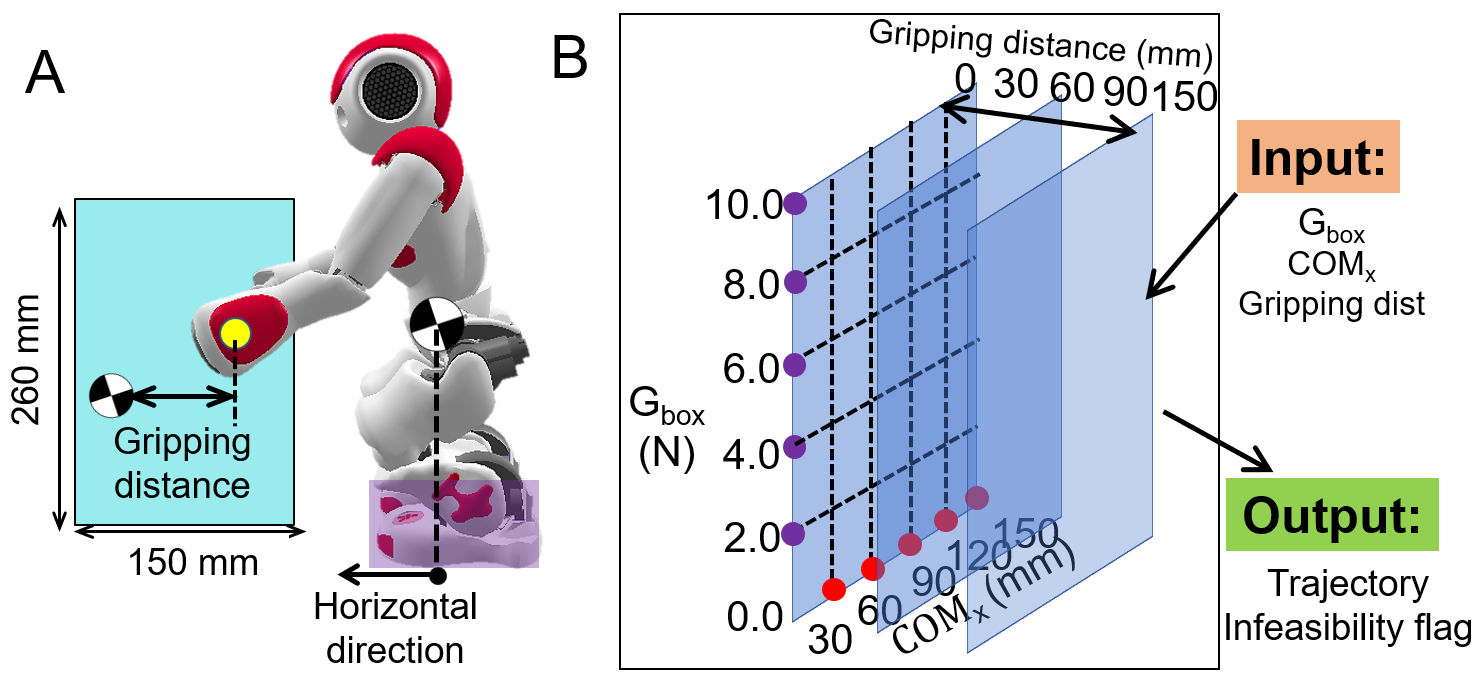}
\caption{A. A diagram showing the definition of horizontal direction relative to the robot and gripping distance. B. A range of feasible physical parameters of the box is discretized to construct the trajectory table and the inputs and outputs of the trajectory table}
\label{Fig.table}
\end{figure} 

\section{Trajectory Table Construction}
A method for constructing the trajectory table for lifting feasibility evaluation is introduced, which corresponds to the dashed box on the left side of Fig. 2. The trajectory table stores either valid motion trajectories or infeasibility flags related to different combinations of the discretized physical parameters of the box. 
\subsection{Parameter Space Discretization}
The feasibility of the robot's lifting motion is dependent upon the physical properties of the box, including its weight, location of COM and surface friction. To determine the crucial parameters corresponding to the box for discretization and later indexing in the trajectory table, we simplify the system in the following ways. First, to reduce the modeling complexity, we assume that the robot moves quasi-statically and that the COM of the box is located in the robot's sagittal plane. Second, the robot uses a fixed nominal gripping force to hold the box. In addition, we added the constraint that the box orientation does not change along the trajectory, which, in practice, prevents objects in the box from slipping. Taking these into account, the stability of the system and the maximum torque required to maintain static motion only depend on the configuration of the robot and the COM location of the box in the horizontal direction (Fig. 3A). We further define the gripping distance as the distance between the COM of the box and the gripping location in the horizontal direction (Fig. 3A). This distance ensures that the gravitational torque of the box ($G_{\text{box}}\times$gripping distance) can be balanced by the torques generated by robot's hands under nominal gripping forces, and it reflects the friction properties of the surface of the box. Accordingly, we choose to discretize the weight, location of COM in the horizontal direction and the gripping distance corresponding to the box within their allowable ranges (Fig. 3B). In our case, a box is a 260 mm $\times$ 150 mm $\times$ 140 mm cuboidal object. The COM in horizontal direction and the gripping distance are bounded between 0 mm and 150 mm, with 30 mm increments. The weight of the box is bounded between 0 N and 10 N with 2 N increments, which are based on the pre-simulated results of whether the robot can lift up the box or not.

\subsection{Trajectory Generation}\label{TG}
There are many studies focusing on generating motion trajectories for robotic manipulators and humanoid robots. Shiller et al. introduced an optimal control frame work in constrained environments for robotic manipulators \cite{shiller1985optimal}. Nakamura et al. developed a global optimal redundancy control strategy for manipulators \cite{nakamura1987optimal}. Kuindersma et al. applied trajectory optimization for dynamic motion planning of the Atlas robot \cite{kuindersma2016optimization}.
Kuffner et al. proposed a whole body planning approach for humanoid robots using a sampling based method together with a dynamic balance filter \cite{kuffner2002dynamically}. A differential dynamic programming framework was also introduced for trajectory generation for the humanoid robots \cite{tassa2014control}. Burget et al. proposed an improved RRT sampling based method for whole body static motion planning of the NAO robot \cite{burget2013whole}. Considering the complex path constraints of our system, we adopt the trajectory optimization approach in the optimal control framework to generate trajectories for each combination of the discretized indices. The formulation of a general trajectory optimization problem is given by:
\begin{align}
\underset{x,u}{\text{minimize}} \quad & J = \phi(x(t_{f}),t_{f})+\int_{t_{0}}^{t_{f}}L(x(t),u(t),t)dt\\ 
\textrm{subject to} \quad & \dot{x}(t) = f(x(t),u(t),t)\\
&c(x(t),u(t),t)	\leqslant 0\\
&\psi(x(t_{f}),t_{f}) = 0,
\end{align}
in which $x(t)$ and $u(t)$ in (1) are respectively the state space vector and the control vector. $J$ in (1) is the cost function, which includes a terminal state cost $\phi(x(t_{f}),t_{f})$ along with a path dependent penalization term $L(x(t),u(t),t)$. The system states are subject to the dynamics in (2), control and state constraints along the path in (3), and the terminal constraints at the final time in (4).

In our case, we formulate the trajectory optimization into a direct collocation problem \cite{tsang1975optimal}. This method discretizes a trajectory into separate time intervals and represents differential constraints as algebraic equations in each interval. Then the problem is solved using nonlinear programming. Specifically, we define our system states using joint angles $\mathbf{q} = [q_{1},q_{2}, \hdots, q_{m}]$. The discretized optimization problem for robot's quasi-static motion is given by
\begin{align}
\underset{\mathbf{q,\dot{q}}}{\text{minimize}} \quad & J = \phi(\mathbf{q}[N]) + \sum_{k=1}^{N}L_{k}(\mathbf{q}[k])\\
\textrm{subject to} \quad 
&\dot{\mathbf{q}}[k] = \mathbf{u}[k]\\
&\text{COP}_{x}[k] = \text{C}(\mathbf{q}[k]) \in \mathbf{S}\\
&\mathbf{q_{\text{min}}}< \mathbf{q}[k] <\mathbf{q_{\text{max}}}\\
&|\mathbf{\tau}[k]| = |\text{T}(\mathbf{q}[k])| <\mathbf{\tau_{\text{max}}}\\
&|\mathbf{u}[k]| <\mathbf{u_{\text{max}}}\\
&\text{Kinematic constraints}
\end{align}
The cost function $J$ in (5) includes a terminal cost
$$
\phi(\mathbf{q}) = \frac{1}{2}(\mathbf{q}[N]-\mathbf{q}_{d})^{T}P_{f}(\mathbf{q}[N]-\mathbf{q}_{d}),
$$
which minimizes the difference between the robot's final and desired states and a cost along the path
$$
L_{k} = ||\mathbf{q}_{d}-\mathbf{q}[k]||^{2}_{Q_{q}} + ||\mathbf{\tau}[k]||^{2}_{Q_{t}} + ||\text{COM}_{x}[k]||^{2}_{Q_{c}},
$$
which 1) drives the robot to its final configuration, 2) maximizes the stability of the robot and 3) minimizes the joint torques of the robot with respect to the optimization variables. Since the robot moves quasi-statically, the control inputs for each time interval can be simplified as constant values and the state transition is defined by (6). Stability constraint in (7) confines the center of pressure (COP) of the robot to lie inside the support polygon $\mathbf{S}$ defined by the convex hull containing the robot's feet (Fig. 3A). The joint angle and torque constraints in (8) and (9) bound the operational angles and torques. A control boundary constraint in (10) bounds the magnitude of the control variables in each interval in order to preserve the quasi-static assumption. The kinematic constraints in (11) ensure that the hand positions and orientations satisfy the gripping requirements and prevent the collision between the box and the robot. The symmetry of the robot in the sagittal plane allows the collision constraints to be simplified by approximating the torso and leg links via their enclosing 2D ellipses. We impose the constraints that the box does not penetrate these ellipses.

The initial and goal configurations of the robot for box lifting are generated by two constrained optimization processes. The initial configuration is generated by solving
\begin{align*}
\underset{\mathbf{q}}{\text{minimize}} \quad & \text{w}_{1}\text{COM}_{x}^{2}+|\mathbf{\tau}|^{T}\mathbf{w_{2}}|\mathbf{\tau}|+ \text{w}_{3}\tau_{\text{hand}}^{2}\\
\textrm{subject to} \quad
&\text{COP}_{x} \in \mathbf{S}\\
&\mathbf{q}_{\text{min}}<\mathbf{q}<\mathbf{q}_{\text{max}}\\
&|\mathbf{\tau}|<\mathbf{\tau}_{\text{max}}\\
&|\mathbf{p}_{\text{hand}x}- \text{COM}_{x}| < \text{gripping distance} \\
&\text{Kinematic constraints},
\end{align*}
which maximizes the stability of the robot while minimizing the joint and hand torques during the initial lifting process, subject to some of the mentioned trajectory optimization constraints. An extra constraint requiring the initial gripping distance to be smaller than the estimated gripping distance is introduced, which guarantees sufficient frictional torque at the gripping position under nominal gripping forces to prevent the box from rotating during the lifting process. The optimization for generating goal configuration is given by:
\begin{align*}
\underset{\mathbf{q}}{\text{minimize}}\quad & \text{w}_{4}\text{COM}_{x}^{2}+|\mathbf{\tau}|^{T}\mathbf{w_{5}}|\mathbf{\tau}|\\
\textrm{subject to}  \quad 
&\text{COM}_{x} \in \mathbf{S}\\
&\mathbf{q}_{\text{min}}<\mathbf{q}<\mathbf{q}_{\text{max}}\\
&|\mathbf{q}_{l}|<\sigma\\
&|\mathbf{\tau}|<\mathbf{\tau}_{\text{max}}\\
&\theta_{t_{f}} = \theta_{0}\\
&\text{Kinematics constraints}.
\end{align*}
In calculating the robot's goal configuration, further constraints are imposed. The legs' final joint angles $\mathbf{q}_{l}$ are bounded by a threshold to form a normal standing posture and the final orientation of the box $\theta_{t_{f}}$ at the goal configuration is set equal to that of its initial configuration $\theta_{0}$. 

The trajectory optimization process is implemented in an open source optimal control solver ACADO \cite{Houska2011a}.
\subsection{Table Construction}
For each combination of the discretized indices, if there exist a pair of valid initial and desired configurations and the trajectory optimization process converges within a given number of iterations, the task is considered to be feasible and the trajectory is saved in the table. Otherwise, an infeasibility flag is saved.
\section{Physical Property Identification}
\begin{figure}[t!]
\centering
\includegraphics[width=1\linewidth]{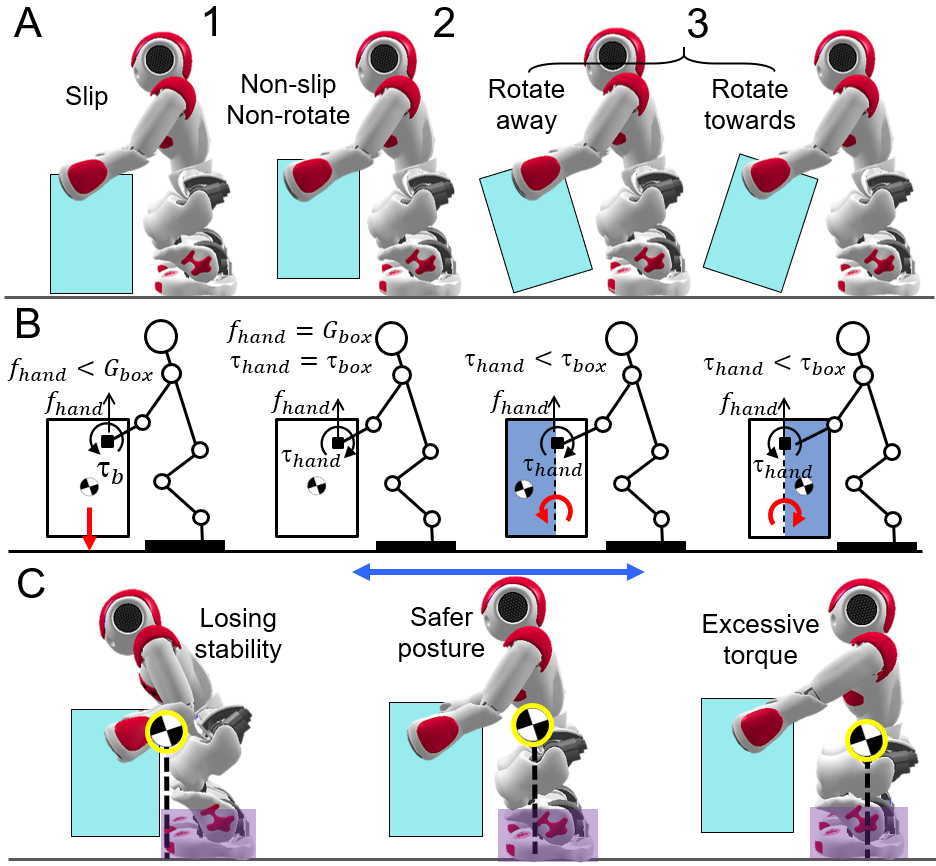}
\caption{A. Three possible motions of the box relative to the gripping spot when the box is slightly lifted off the ground. B. The force torque analysis according to each condition in A. black arrows represent the forces and torques generated by the hands of the robot to prevent the box from its relative motion and red arrows represent the relative motions of the box due to gravitational force. C. Example of a posture which loses stability (left) or suffers from excessive torque (right).}
\label{Fig.reason}
\end{figure} 
A physical-interaction-based planning algorithm is developed for the robot to identify the weight, location of COM and gripping distance corresponding to the box. This algorithm first searches for a robust gripping position so that the robot can lift the box slightly off the ground, in the sense that the box will not slip or rotate, through the robot's visual feedback. During this process, this method ensures the stability of the robot and limits the electric current (torque) of the actuators by optimizing the robot's posture using the feedback from its foot and electric current sensors (see \ref{exp}). If the robot successfully finds a robust gripping position to lift the box slightly off the ground, it records this gripping distance and estimates the weight, the location of COM in the horizontal direction of the box. Otherwise, the lifting task is flagged as infeasible. The parameter identification process refers to the middle part of Fig. 2. and the detailed planning algorithm is shown in \textbf{Algorithm 1}.
\subsection{Robust Gripping Distance Searching}
Assuming that the robot is able to lift up the box slightly off the ground with nominal gripping forces while maintaining stability and limiting its joint torques, three possible motions of the box relative to the gripping spot may take place resulting from different frictional force and torque interactions between the hands of the robot and the surfaces of the box (Fig. 4A, B). Here we denote the frictional forces and torques generated by the hands of the robot as $f_{\text{hand}}$ and $\tau_{\text{hand}}$, the gravitational force of the box as $G_{\text{box}}$ and the gravitational torque of the box to be balanced as $\tau_{\text{box}}$. The three outcomes are listed as follows:\\*
\textbf{1.\:Slip}: The box slips off the hands and falls onto the ground due to insufficient frictional forces $f_{\text{hand}}$ acting on the box. In this case, $f_{\text{hand}}$ is less than $G_{\text{box}}$ and the box cannot be lifted regardless of the magnitude of $\tau_{\text{hand}}$ (Fig. 4A, B1).
\\*
\textbf{2.\:Non-slip/Non-rotate}: The box is held firmly by the hands without slipping or rotating around the gripping spot. In this case, $f_{\text{hand}} = G_{\text{box}}$ and $\tau_{\text{hand}} = \tau_{\text{box}}$. If the gripping distance is smaller than or equal to the current measurement, it is assumed that there are sufficient frictional forces and torques to lift up the box under nominal gripping forces (Fig. 4A, B2).
\\*
\textbf{3.\:Rotate}: The box is lifted up but rotates either towards the robot or away from it, as shown in Fig. 4A (3). In this case, $\tau_{\text{box}} > \tau_{\text{hand}}$ and $f_{\text{hand}} \leqslant G_{\text{box}}$ (Fig. 4B, 3). If the box rotates away from the robot, a potential robust gripping distance may be located in the direction pointing away from the robot  (Fig. 4B, 3, left, blue area). Conversely, if the box rotates towards the robot, a potential robust gripping distance position may be located in the direction pointing towards the robot (Fig. 4B, 3, right, blue area).

By quantifying different possible configurations of the box during lifting attempts, the robot is capable of 1) discerning whether there is an available gripping position, 2) determining whether to start the box parameter estimation, 3) adjusting its gripping position by observing the change of the orientation of the box via its visual feedback (see \ref{exp}). 

\subsection{Posture Adjustments}
During the initial lifting attempt, the robot may lose stability and the torques of the actuators may exceed their limits. We developed a strategy to handle the above situations by adjusting robot's posture through its sensory feedback.   \\*
\textbf{1. Losing stability}:
once the robot's COP is detected to pass a stability threshold (Fig. 4C, left), the robot first releases the box and then adjusts itself to a new lifting posture, which is generated by solving a constrained optimization problem. This problem maximizes the stability of the robot while imposing constraints
to confine the COM of the new posture to be at a certain horizontal distance closer to the center of the support polygon than that of the previous posture. The gripping distance of the two postures remain the same.
\\*
\textbf{2. Excessive torques}:
same as the above approach, once the torques of certain actuators are detected to exceed their maximum limits (Fig. 4C, right), the robot adjusts its posture by solving another constrained optimization problem, which applies additional constraints to limit the torques of those joints in the new posture, affected by the gravitational forces of the robot's joints and links in dependent to the box, to be at a certain magnitude smaller than that of the previous posture while maintaining the same gripping distance.

\subsection{Parameter Identification}
If the box is successfully lifted off the ground, the total weight combining the robot and the box equals to the ground reaction force $N$, which is measured by the force sensors (see section \ref{exp}). Then the weight of the box can be estimated as
$$
G_{\text{box}} = N - G_{\text{robot}}.
$$
The COM location of the box in the horizontal direction of the spatial frame, $p_{\text{box}}$, can be estimated by solving
$$
\text{COP}_{x} = \frac{\sum_{i=1}^{n}m_{i}p_{i}+m_{\text{box}}p_{\text{box}}}{\sum_{i=1}^{n}m_{i}+m_{\text{box}}},
$$
where $\text{COP}_{x}$ is the COP of the robot in the horizontal direction relative to the spatial frame, which can be measured by the force sensors (see section \ref{exp}). $m_{i}$ and $p_{i}$ are the mass and the COM location of robot's i-th link, or joint, in the horizontal direction of the spatial frame, which can be obtained by solving forward kinematics of the robot using the encoder data. Eventually, $\text{COP}_{x}$ can be transferred to the body frame of the box by obtaining the relative transformation between the box and the robot through robot's vision feedback. The desired parameters are estimated using the averaged data after robot reaches to its steady state.
\begin{algorithm}[t!] \label{ALG}
\caption{Parameter Identification}
\begin{algorithmic}
\STATE Lifting posture initialization \\
$\text{Attempt}\leftarrow$ 0 \\ 
\WHILE{$\text{Attempt} \leqslant N$}
\STATE $\text{Attempt} = \text{Attempt} + 1 $
\STATE Start lifting and monitoring the following cases
\IF{$\text{COP}_{x} \leqslant \textbf{S}_{t}$ and $|\mathbf{\tau}| \leqslant \mathbf{\tau}_{t}$ (posture is safe)}
\IF{$h_{\text{box}} < h_{\text{t}}$ (box slips) and robot reaches goal}
\STATE$\textbf{return}$ Infeasibility flag
\STATE$\textbf{break}$
\ELSIF{$h_{\text{box}} \geqslant h_{\text{t}}$ and $|\theta_{\text{box}}| \leqslant \theta_{t}$ (box is lifted)} 
\STATE $\textbf{return}$ $\text{gripping distance}, G_{\text{box}}, \text{COM}_{\text{box}}$
\STATE$\textbf{break}$
\ELSIF{$\theta_{\text{box}} > \theta_{t}$ (box rotates away)}
\STATE Adjust posture s.t. $\text{grip}_{x}$ = $\text{grip}_{x}$ + d (grip further) 
\ELSIF{$\theta_{\text{box}} < -\theta_{t}$ (box rotates towards)}
\STATE Adjust posture s.t. $\text{grip}_{x}$ = $\text{grip}_{x}$ - d (grip closer)
\ENDIF
\ELSIF{$\text{COP}_{x} > \textbf{S}_{t}$ (losing stability)} 
\STATE Adjust posture s.t. $\text{COP}_{x}^{\text{i}} - \text{COP}_{x}^{\text{i+1}} > \sigma_{c}$, while keeping the same gripping distance 
\ELSIF{$\quad |\mathbf{\tau}| > \mathbf{\tau}_{t}$ (excessive torque)}
\STATE Adjust posture s.t. $\tau_{\text{j}}^{\text{i}} - \tau_{\text{j}}^{\text{i+1}} > \sigma_{\tau}$, while keeping the same gripping distance 
\ENDIF
\ENDWHILE
\end{algorithmic}
\end{algorithm}
% \begin{algorithm}[t!] \label{ALG}
% \caption{Parameter Identification}
% \begin{algorithmic}
% \STATE $\text{Initializing lifting posture}$ \\
% $\text{Attempt}\leftarrow$ 0 \\ 
% \WHILE{$\text{Attempt} \leqslant N$}
% \STATE  Lifting attempt with nominal gripping forces
% \IF{$\text{COP}_{x} < \textbf{S}_{\text{max}} \quad \text{and} \quad |\mathbf{\tau}| < \mathbf{\tau}_{max}$}
% \IF{$|\theta| < \delta$ ($\theta$: the tilting angle of the box)} 
% \STATE $\textbf{return}$ $\text{gripping distance}, G_{\text{box}}, \text{COM}_{\text{box}}$
% \ELSIF{$|\theta| > \delta$}
% \IF{$\theta > 0$ (box rotates away)}
% \STATE Optimize posture s.t. $\text{grip}_{x}$ = $\text{grip}_{x}$ + d 
% \ELSIF{$\theta < 0$ (box rotates towards)}
% \STATE Optimize posture s.t. $\text{grip}_{x}$ = $\text{grip}_{x}$ - d 
% \ENDIF
% \ELSIF{h $< \text{h}_{\text{min}}$ (box slips onto the ground)}
% \STATE$\textbf{return}$ Infeasibility flag
% \STATE$\textbf{break}$
% \ENDIF
% \ELSIF{$\text{COP}_{x} > \textbf{S}_{\text{max}}$ (losing stability)} 
% \STATE Optimize posture s.t. $\text{COP}_{x}^{\text{i}} - \text{COP}_{x}^{\text{i+1}} > \sigma_{1}$, while keeping gripping distance 
% \ELSIF{$\quad |\mathbf{\tau}| > \mathbf{\tau}_{\text{max}}$ (excessive torque)}
% \STATE Optimize posture s.t. $\tau_{\text{j}}^{\text{i}} - \tau_{\text{j}}^{\text{i+1}} > \sigma_{2}$, while keeping gripping distance 
% \ENDIF
% \ENDWHILE
% \end{algorithmic}
% \end{algorithm}
\section{Table Mapping}
Once the gripping distance and the physical properties of the box are estimated, they are matched with the existing discretized indices of the trajectory table. The $\text{COM}_{x}$ and weight indices are chosen to be the closest values in the table that are larger than the estimated values since the task remains feasible for the robot in less extreme planning paradigms. For the gripping distance, the index is chosen to be the closest smaller recorded value to ensure sufficient torque, which prevents the box from rotating. If the trajectory corresponding to the matched indices exists, the task is marked as feasible and the robot receives the trajectory. Otherwise, the task is flagged as infeasible. The table mapping process refers to the dashed box on the right side of Fig. 2.
\section{Experiments} \label{exp}
\subsection{Experimental Setup}
\begin{figure}[t!]
\centering
\includegraphics[width=0.95\linewidth]{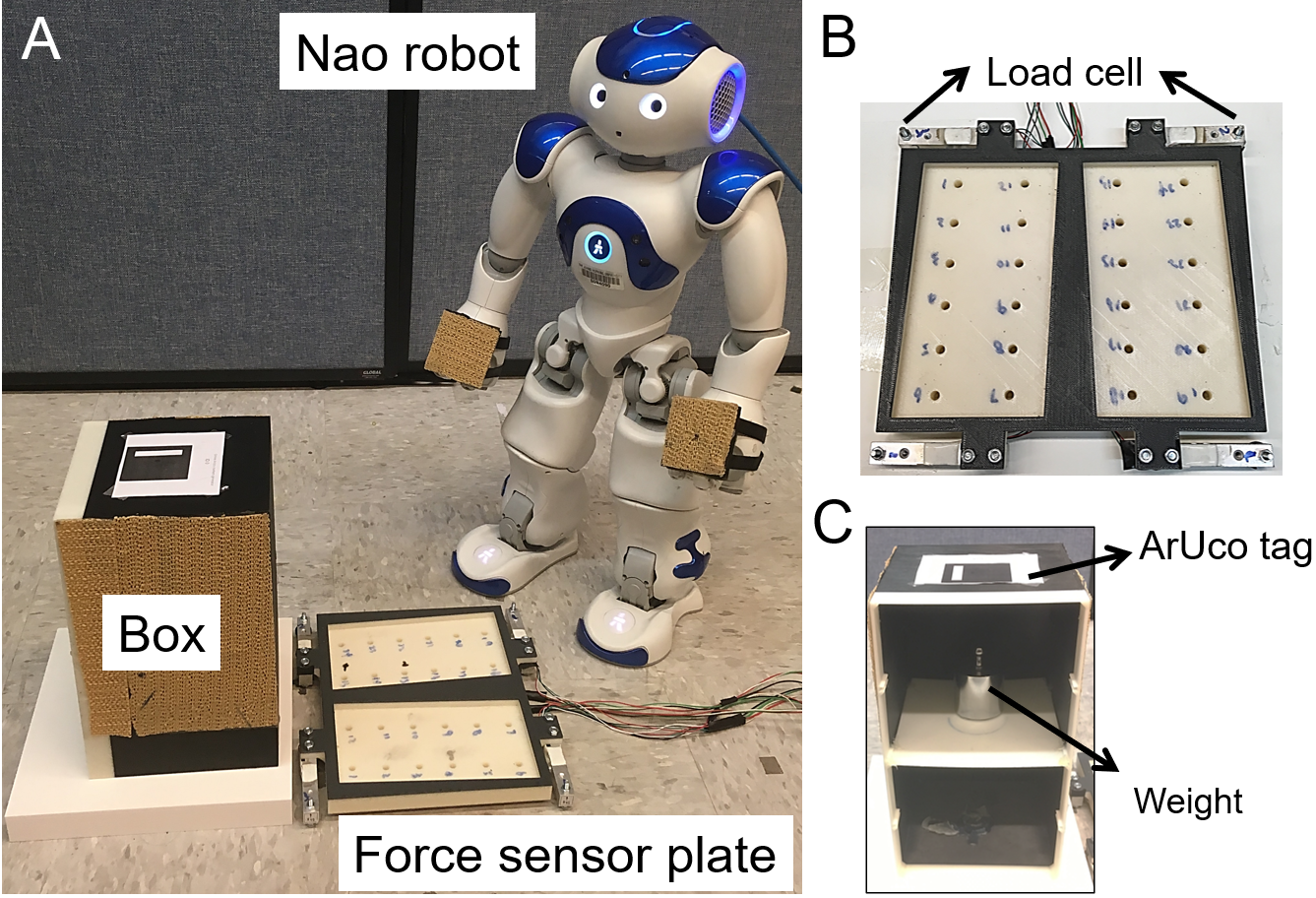}
\caption{A. Experimental setup includes a Nao robot, a force sensor plate and a box. B. The force sensor plate consists of 4 load cells. C. The box is designed with one side open to contain different weights and an ArUco tag is attached to the top of the box.}
\label{Fig.experiment}
\end{figure}
We performed experiments to present our method. The experimental setup is shown in Fig. 5A, which includes a Nao robot H25 V5, a force sensor plate and a box. Two frictional pads were attached to the hands of the robot to increase the gripping friction. The force sensor plate, constructed with four load cells (Fig. 5B), measures the ground normal force and the COP of the robot, which significantly improved the accuracy of measurement compared to the robot's built-in foot pressure sensors. The ground reaction force $N = \sum_{i=1}^{4}F_{i}$ and the COP in the horizontal direction of the robot $\text{COP}_{x} = \frac{\sum_{i=1}^{4}F_{i}p_{i}}{\sum_{i=1}^{4}F_{i}}$. $F_{i}$ is the force reading of the $i$th load cell and $p_{i}$ is the horizontal distance of the measuring point of the $i$th load cell relative to the spatial frame attached to the foot of the robot. The box was designed with one open side (Fig. 5C) so that external weights 
\begin{figure}[tp]
\centering
\includegraphics[width=1\linewidth]{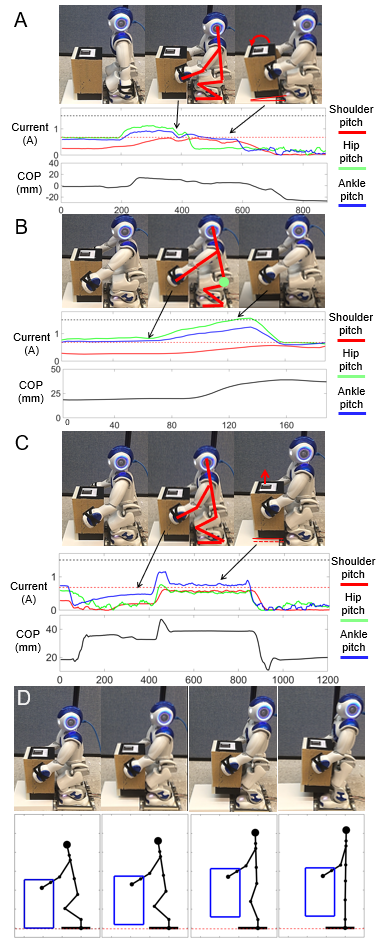}
\caption{The Nao slightly lifts up the box to estimate its parameters through three attempts and then executes the trajectory obtained from the trajectory table corresponding to the mapped indices. A, B, C show the snapshots and sensor readings for the first, second and third lifting attempts. The red dashed line is the electric current limit of the actuator of shoulder pitch and the black dashed line is the electric current limit of the actuators of the hip pitch and ankle pitch. D. The snapshots of the robot's whole body lifting motion and its corresponding simulation result of the stored trajectory.}
\label{Fig.reason}
\end{figure} 
could be added to change its weight and COM location. In the experiments, the robot used its bottom camera to detect the ArUco \cite{garrido2014automatic} tag attached to the top of the box (Fig. 5C) and acquire the position and orientation of the box relative to the spatial frame attached to the robot. The motor encoder data was used to calculate the COM position of the robot and the electric current sensor data was used to monitor the maximum torques of the actuators. 

\subsection{Demo}
A demo of the robot reasoning about the feasibility of lifting a 0.85 kg box ($\approx 0.16 \times G_{\text{Robot}}$) is shown in Fig. 6. The robot first initialized its posture to lift the box slightly off the ground in the vertical direction (Fig. 6A). During the first lifting attempt, the change in the box's orientation was detected from robot's visual feedback (Fig. 6A, right). Then the robot released the box and subsequently adjusted its posture to grip further away from itself (Fig. 6B). During the second lifting attempt, the electric current of the hip pitch actuator exceeded the safety limit (Fig. 6B, green curve. For simplicity, only three relatively higher electric current values are shown). The robot released the box again and adjusted its posture by reducing its hip pitch torque while fixing the gripping distance. In the third attempt, the robot successfully lifted the box slightly off the ground and estimated its parameters (Fig. 6C). Finally, one combination of indices in the trajectory table was mapped to the estimated parameters and the task was flagged as feasible since there existed a valid trajectory. Fig. 6D shows that the robot successfully lifted up the box along the stored trajectory in the table and the motion of the robot matched with the simulation result. 

\section{Discussion and future work}
This paper proposed a three-step method for humanoid robots to reason about the feasibility of lifting a heavy box with a priori unknown physical properties. A lifting trajectory table is first constructed with discretized indices corresponding to a range of feasible physical parameters of the box and the gripping distance. A physical-interaction based search algorithm is proposed to estimate the actual parameter values before lifting, while maintaining the stability and safety of the robot. The estimated parameter values are then matched with the indices in the table, which allows the robot to make a judgement about the feasibility of the task by querying the existence of a valid trajectory corresponding to the matched indices. 

Compared to traditional methods, our approach can quickly provide a feasibility reasoning result and a usable trajectory for a lifting task by circumventing a computationally intensive optimization process prior to each specific lifting process. The obtained trajectory is a near-optimal discretized solution rather than a continuous globally optimal one. But such a solution preserves the safety and stability of the robot, which can be applied in practical scenarios. In addition, it is natural to extend our approach to more complex conditions of the lifting task, such as dynamic lifting and considering a priori unknown geometry of the box, which simply require the expansion of the parameter space of the trajectory table.

The future work includes implementing our approach on different objects for lifting tasks, and developing state estimation and feedback control algorithms to deal with the uncertain conditions of the object during the lifting process.

% \section{Conclusions}
% This paper proposed a three-step method for humanoid robots to reason about the feasibility of lifting a heavy box with unknown physical properties. A lifting trajectory table is first constructed with discretized indices corresponding to the physical parameters of the box and the gripping distance. A physical-interaction based searching algorithm is proposed to identify these parameters, while maintaining the stability and safety of the robot. The estimated parameters are then matched with the indices in the table, which allows the robot to make a judgement on the feasibility of the task by querying the existence of a valid trajectory corresponding to the matched indices. Our method is capable of efficiently evaluating the feasibility of a lifting task through simple interactions between the robot and the box, while simultaneously obtaining the desired motion trajectory. We successfully demonstrated the proposed method using a NAO humanoid robot. Our method can be applied to other challenging tasks for other robots.

%%%%%%%%%%%%%%%%%%%%%%%%%%%%%%%%%%%%%%%%%%%%%%%%%%%%%%%%%%%%%%%%%%%%%%%%%%%%%%%%

%%%%%%%%%%%%%%%%%%%%%%%%%%%%%%%%%%%%%%%%%%%%%%%%%%%%%%%%%%%%%%%%%%%%%%%%%%%%%%%%

%%%%%%%%%%%%%%%%%%%%%%%%%%%%%%%%%%%%%%%%%%%%%%%%%%%%%%%%%%%%%%%%%%%%%%%%%%%%%%%%

\section*{ACKNOWLEDGMENT}

The authors thank Ruoyu Lin, Karen Poblete Rodriguez, Qiangqiang Zhao, Thomas Mitchel, Jim Seob Kim, Timothy Ma and Marin Kobilarov for helpful discussion. This work was performed under National Science Foundation grant IIS1619050 and Office of Naval Research Award N00014-17-1-2142. The ideas expressed in this paper are solely those of the authors.

%%%%%%%%%%%%%%%%%%%%%%%%%%%%%%%%%%%%%%%%%%%%%%%%%%%%%%%%%%%%%%%%%%%%%%%%%%%%%%%%

\bibliographystyle{IEEEtran}
\bibliography{IROS.bib}
\end{document}